\pdfoutput=1

\documentclass[11pt]{article}

\usepackage{ACL2023}

\usepackage{times}
\usepackage{latexsym}

\usepackage[T1]{fontenc}

\usepackage[utf8]{inputenc}

\usepackage{microtype}

\title{\model: A Unified Framework for Relational Structure Extraction} 

\author{I-Hung Hsu\thanks{\; The authors contribute equally.}$^{\;\;\dagger}$ \ \ \ Kuan-Hao Huang\footnotemark[1]$^{\;\;\ddagger}$ \ \ \ Shuning Zhang$^{\diamond}$ \ \ \  Wenxin Cheng$^{\ddagger}$ \\{\bf Premkumar Natarajan$^{\dagger}$} \ \ \ {\bf Kai-Wei Chang$^{\ddagger}$ \ \ \ Nanyun Peng$^{\ddagger}$} \vspace{0.2em}\\
$^{\dagger}$Information Science Institute, University of Southern California \\
$^{\ddagger}$Computer Science Department, University of California, Los Angeles \\ 
$^{\diamond}$Computer Science and Technology Department, Tsinghua University \vspace{0.2em}\\
\texttt{\{ihunghsu, pnataraj\}@isi.edu}, \\
\texttt{\{khhuang, kwchang, violetpeng\}@cs.ucla.edu} \\
\texttt{zhang-sn19@mails.tsinghua.edu.cn}, $\;$ \texttt{wenxin0319@ucla.edu} \\
}

\usepackage{graphicx}
\usepackage{multirow}
\usepackage{makecell}
\usepackage{booktabs}
\usepackage{enumitem}
\usepackage{amssymb}
\usepackage{xcolor}
\usepackage{subcaption}
\usepackage{threeparttable}
\definecolor{dark-green}{rgb}{0.31, 0.47, 0.26}
\definecolor{dark-red}{rgb}{0.81, 0.09, 0.13}
\usepackage{soul}
\usepackage{amsmath}

\setuldepth{Berlin}
\usepackage{tabularx}
\newcolumntype{x}[1]{>{\arraybackslash\hspace{0pt}}m{#1}}

\usepackage{pifont}
\newcommand{\cmark}{\color{dark-green}{\ding{51}}}
\newcommand{\xmark}{\color{dark-red}{\ding{55}}}
\newcommand{\model}{\textsc{TagPrime}}

\newcommand{\mbf}[1]{\mathbf{#1}}
\newcommand{\mypar}[1]{\vspace{0.3em}\noindent\textbf{#1}}

\usepackage{todonotes}

\begin{document}
\maketitle
\begin{abstract}
Many tasks in natural language processing require the extraction of relationship information for a given condition, such as event argument extraction, relation extraction, and task-oriented semantic parsing. Recent works usually propose sophisticated models for each task independently and pay less attention to the commonality of these tasks and to have a unified framework for all the tasks. In this work, we propose to take a unified view of all these tasks and introduce \model{} to address relational structure extraction problems. \model{} is a sequence tagging model that appends \emph{priming words} about the information of the given condition (such as an event trigger) to the input text. 
With the self-attention mechanism in pre-trained language models, the priming words make the output contextualized representations contain more information about the given condition, and hence become more suitable for extracting specific relationships for the condition. Extensive experiments and analyses on three different tasks that cover ten datasets across five different languages demonstrate the generality and effectiveness of \model{}.
 
\end{abstract}

\section{Introduction}
\label{sec:intro}
There are many tasks in natural language processing (NLP) that require extracting relational structures from texts. 
For example, the event argument extraction task aims to identify event arguments and \textit{their corresponding roles} for a given event trigger~\cite{acl2022xgear, wang-etal-2019-hmeae}. 
In entity relation extraction, the model identifies the tail-entities and head-entities that \textit{forms specific relations}~\cite{DBLP:conf/acl/WeiSWTC20, DBLP:conf/ecai/0002ZSLWWL20}. 
In task-oriented semantic parsing, the model predicts the slots and \textit{their semantic roles} for a given intent in an utterance \cite{Tur10atis,Li21mtop}.
These tasks are beneficial to a wide range of applications, such as dialog systems~\cite{liu2018knowledge}
, question answering~\cite{yasunaga-etal-2021-qa}
, and narrative generation~\cite{chen2019incorporating}.
Prior works usually design models to specifically address each of the tasks~\cite{DBLP:conf/acl/SunGWGJLSD19, DBLP:conf/acl/MiwaB16, han-etal-2019-deep, DBLP:conf/acl/FuLM19, DBLP:conf/icann/ZhangFCLCT18}.
However, less attention is paid to the commonality among these tasks and having a unified framework to deal with them and provide a strong baseline for every task. 

In this work, we take a unified view of these NLP tasks.
We call them relational structure extraction (RSE) tasks and formulate them as a unified task that identifies arguments to a given condition and classifies their relationships.
The condition could be a textual span, such as an event trigger for event argument extraction, or a concept, such as an intent for task-oriented semantic parsing.

\begin{figure*}[t!]
    \centering
    \includegraphics[trim=0cm 0cm 0cm 0cm, clip, width=0.98\textwidth]{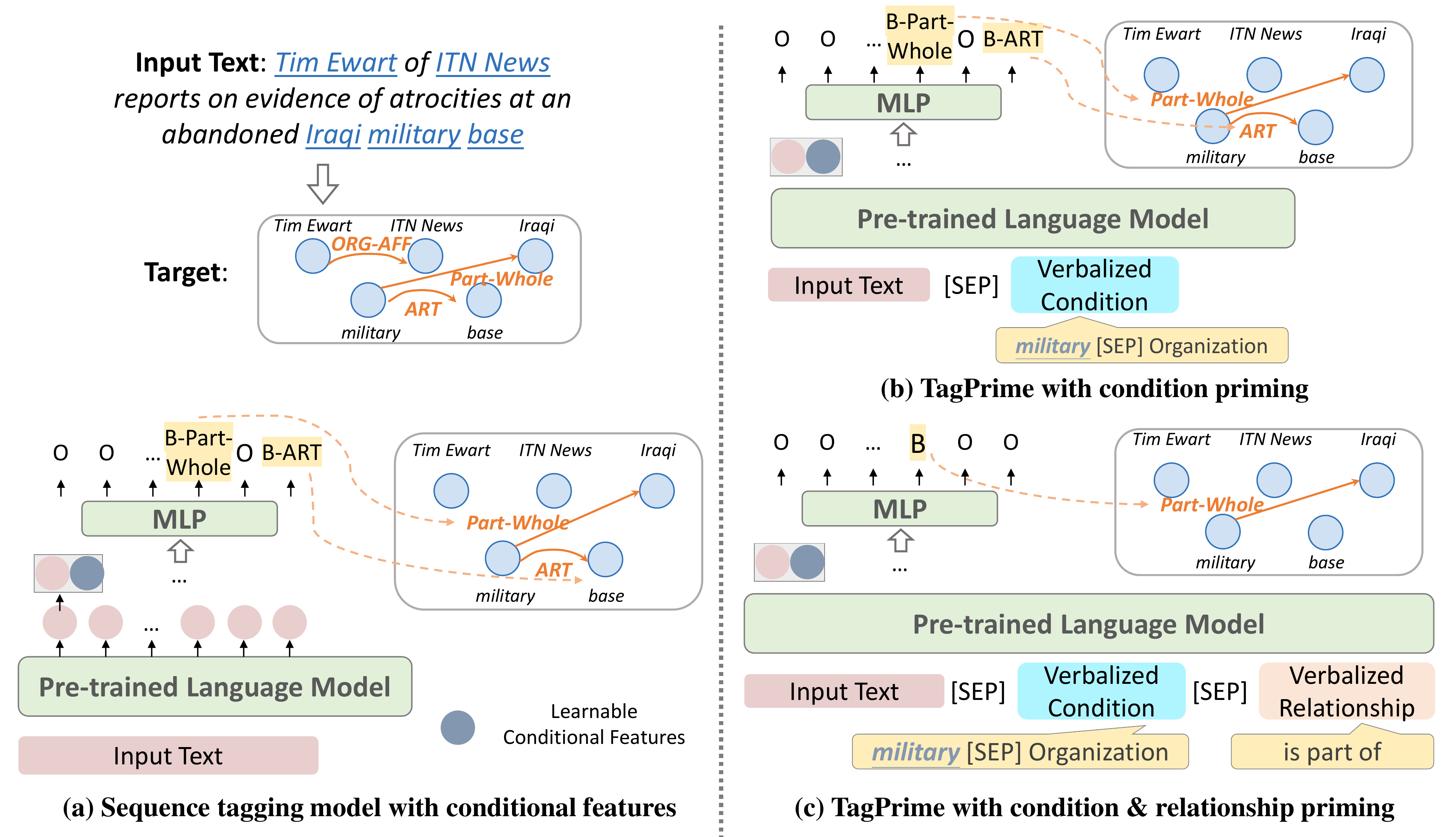}
    \caption{Illustrations of different models running on an instance for relation extraction, where the target is to predict the relations between the named entities.
    (a) \textbf{Sequence tagging model with conditional features}: A conventional sequence tagging model that embeds conditional information by adding learnable features to the output representation from a pre-trained language model. In the shown example, the conditional features contain two parts: one is the token embedding representing the conditional word \emph{``military''}, and the other is an entity type embedding. 
    (b) \textbf{\model{} with condition priming}: The conditional information is further applied to the input sequence to induce the output representation from the pre-trained language model to become condition-aware.
    (c) \textbf{\model{} with condition \& relationship priming}: Our approach that further append the verbalized relationship to \model{} with condition priming model. For this case, the goal of the tagging model is to make predictions specific to the relationship type in the input. We omit CRF layers after MLP layers in this figure for better readability.}
    \label{fig:mthd_overview}
\end{figure*}

We present \model{}, a simple, unified, and strong model, which follows a sequence tagging paradigm with a \emph{priming technique}, which is proposed by \citet{DBLP:journals/corr/abs-2109-12383}. 
\model{} inherits the strength of sequence tagging models to unifiedly address RSE by converting the relational structure into a sequence of predictions by sequentially labeling tokens in the input passage.
\model{} further improves this framework's performance by better incorporating information about the given condition via priming.
Traditional sequence tagging models usually leverage learnable feature embeddings to incorporate information about the given condition before the tags are assigned, as illustrated in Figure~\ref{fig:mthd_overview}(a). With the priming mechanism, \model{} augments the input text with condition-specific contexts, as illustrated in Figure~\ref{fig:mthd_overview}(b) \& (c).
The main merit of the priming technique comes from the nature of the self-attention mechanism in pre-trained language models. Augmenting input text with condition-specific contexts makes the sentence representations \emph{condition-specific} directly. Thus, it unlocks the capability of sequence tagging methods for relational structure extraction better than the commonly used feature embedding approach, as shown in Section~\ref{sec:analysis}.


Our contributions can be summarized as follows. 
(1) We take a unified view of NLP tasks that requires extracting relational structures, including end-to-end event extraction, end-to-end relation extraction, and task-oriented semantic parsing. Then, we present \model{}, a unified sequence tagging model with priming that can serve as a strong baseline to various relational structure extraction problems.
(2) 
Thorough experiments on three different tasks show that \model{} achieves competitive performance than the current state-of-the-art on ten datasets in five different languages. 
(3) We propose a novel efficient approximation to speed up \model{} during inference time without sacrificing too much performance. 


\section{Related Work}
Many natural language processing applications require extracting relational structures, including event extraction, relation extraction, coreference resolution, etc.
The prevalence of these applications makes us hard to exhaustively list them in this short summary, hence, we mainly focus on related works for the applications we experiment on.

\paragraph{Event extraction.}
Early works in event extraction mostly consider a pipelined approach~\cite{nguyen-grishman-2015-event, wang-etal-2019-hmeae, DBLP:conf/acl/YangFQKL19} to deal with event extraction. Some follow-up works argue that pipelined design leads to error propagation issues and hence propose end-to-end approaches to better capture dependencies between each prediction \cite{Lin20oneie, Li13jointbeam, Nguyen16jrnn, Hsu22degree, text2event, huang2021document}. However, recently, some empirical studies \cite{Hsu22degree, Zhong21pure, DBLP:journals/corr/abs-2109-12383} also show that when an abundant amount of data is used to learn representations for each pipelined task, 
it is hard to conclude that joint learning approaches always provide a stronger result. This aligns with our discovery in experiments --- even though we apply a pipelined approach with a simple sequence tagging framework on event extraction, with the help of priming to learn more condition-aware contextualized representation, we can still achieve very strong performance on multiple datasets.

\paragraph{Relation extraction.}
End-to-end relation extraction can usually be solved using two categories of approaches. The first one is to directly perform joint inference on named entities and their relation(s)~\cite{DBLP:conf/acl/ZhengWBHZX17,Wang20tableseq, DBLP:conf/acl/KatiyarC17, DBLP:conf/acl/SunGWGJLSD19, DBLP:conf/acl/MiwaB16, DBLP:conf/acl/FuLM19}. The second category is to perform a pipeline that first extracts named entities, and then performs relation classification~\cite{wu2019enriching,hsu2021discourse,lyu-chen-2021-relation,peng-etal-2020-learning,zhou2021improved,lu2022Summ}, which assumes that both the head-entity and tail-entity are given.
Yet, in our unified formulation for relational structure extraction tasks, we extract tail-entities and their corresponding relation types for a given head-entity, which is more similar to a less frequently studied framework called cascading approaches~\cite{DBLP:conf/acl/WeiSWTC20, DBLP:conf/ecai/0002ZSLWWL20}. Despite being a less popular formulation to deal with end-to-end relation extraction, \model{} presents a strong performance compared to prior studies, showcasing the practicality and effectiveness of our unified formulation.

\paragraph{Task-oriented semantic parsing.}
Task-oriented semantic parsing, which focuses on intent classification and slot filling, has a long history of development \cite{Tur10atis,Gupta18top,Li21mtop, DBLP:conf/icann/ZhangFCLCT18, DBLP:conf/coling/LouvanM20}.
Recently, some more advanced neural network-based approaches have been proposed, such as MLP-mixer~\cite{fusco2022mixer} or sequence-to-sequence formulation~\cite{DBLP:journals/corr/abs-2104-07224}. Among them, JointBERT~\cite{Chen19jointbert}, a sequence-tagging-based model that is trained to jointly predict intent and extract slots, serves as a widely-used baseline due to its simplicity. Our approach benefits from the same simplicity as JointBERT and can further improve its performance.
\section{Method}
\label{sec:method}

We first introduce our view to unify RSE problems and then discuss how \model{} approaches this problem under a unified framework of sequence tagging model with priming.

\subsection{A Unified Formulation of RSE}
\label{sec:background}
Given an input text $\mbf x = [x_1, x_2, ..., x_n]$ and a condition~$c$,
The RSE tasks identify a list of spans $\mbf s^c = [s^c_1, s^c_2, ..., s^c_l]$ and their corresponding relationships or attributes $\mbf r^c = [r^c_1, r^c_2, ..., r^c_l]$ towards the condition $c$, where $r^c_i \in \mathcal{A}$ and $\mathcal{A}$ is the set of all possible relationships or attributes.
Many NLP tasks can be formulated as an RSE task. We showcase how this formulation can be applied to event extraction, entity relation extraction, and task-oriented semantic parsing below.

\paragraph{End-to-end event extraction.}
End-to-end event extraction aims to extract events from given texts~\cite{DBLP:conf/emnlp/MaWABA20, Hsu22degree, DBLP:conf/acl/YangFQKL19}.
An event contains a trigger, which is the textual span that best represents the occurrence of an event, and several arguments, which are the participants involved in the event with different argument roles. 
We consider a pipeline solution --- after the event triggers are identified, an argument extraction model extracts the event arguments and their corresponding roles for each given event trigger. Under the RSE formulation, the condition $c$ is the given event trigger, and the target spans $\mbf s^c$ and the relationships $\mbf r^c$ are the arguments and their argument roles, respectively. 

\paragraph{End-to-end relation extraction.}
Relation extraction identifies entities and their relations from texts, and it is usually solved by pipeline approaches --- first extracting named entities and then predicting relations for each entity-pair \cite{wu2019enriching, Zhong21pure}.
Under the new formulation, an RSE model is used to predict \emph{tail-entities} and the relations for each extracted named entity that serves as the \emph{head-entity}. 
For example, in Figure~\ref{fig:mthd_overview}(b), we extract the tail-entities (\emph{``Iraqi''} and \emph{``base''}) and their relation (\emph{``Part-Whole''} and \emph{``ART''}) for the head-entity, \emph{``military''}.
In this way, each given head-entity is the condition $c$, and the extracted tail-entities are $\mbf s^c$, with relations, $\mbf r^c$.

\paragraph{Task-oriented semantic parsing.}
Task-oriented semantic parsing aims to classify the intent and parse the semantic slots in an utterance (to a task-oriented dialog system)~\cite{Li21mtop, Gupta18top}.
To fit into our formulation, we first predict the intent and then use a \emph{relational structure extraction} model to predict the slots ($\mbf s^c$) as well as their semantic roles ($\mbf r^c$) for the given intent ($c$).

\subsection{Sequence Tagging Model for RSE}
\label{sec:basic_tagging}
We hereby introduce the typical way of applying a sequence tagging model to unifiedly solve relational structure extraction.
The goal of our sequence tagging model for relational structure extraction is to predict the BIO-tag sequence $\mbf y=[y_1, y_2, ..., y_n]$, where each $y_i$ is the corresponding tag for each token $x_i$ in the input text. 
The BIO-tag sequence can then be decoded to represent the extracted spans $\mbf s^c$ (and their relationships $\mbf r^c$).

Specifically, given an input text, we obtain the contextualized representation $z_i$ for each token $x_i$ by passing the passage to a pre-trained language model.\footnote{If a token $x_i$ is split into multiple word pieces, we use the average embeddings of all its word pieces to be $z_i$, following the practice of \citet{Lin20oneie}.}
To embed the information of the condition~$c$, one commonly-used technique is to add conditional features to $z_i$~\cite{DBLP:conf/emnlp/MaWABA20, DBLP:conf/acl/WeiSWTC20, DBLP:conf/acl/YangFQKL19, DBLP:conf/ecai/0002ZSLWWL20}, as shown in Figure~\ref{fig:mthd_overview}(a).
For example, in \citet{DBLP:conf/emnlp/MaWABA20}, they use a token embedding of the given event trigger word and a \emph{learnable} event type feature as the conditional features for the task of event argument extraction.
In such case, the feature of $c$ will contain the contextualized word representation $z_j$, if $x_j$ is the token that represents the given condition, i.e., event trigger. 
In our experimental setup, if the given condition can be represented as an input span, we will include the span embeddings as the conditional features together with the type embeddings, such as the cases for event extraction and relation extraction. If the condition is only a concept, such as the task-oriented semantic parsing case, the conditional features will only contain type embeddings.
 Augmented with these conditional features, the final representation for token $x_i$ is further fed into multi-layer perceptrons and a conditional random field (CRF) layer~\cite{DBLP:conf/icml/LaffertyMP01} to predict the BIO-tag sequence~$\mbf y$, as illustrated in Figure~\ref{fig:mthd_overview}(a).

\subsection{\model{}}
\label{sec:priming}
\model{} follows the sequence tagging paradigm but utilizes the priming technique for better leverage information about the input condition.

\paragraph{Condition Priming.}
Motivated by previous work \cite{DBLP:journals/corr/abs-2109-12383}, we consider priming to inject the information of the condition $c$ to further improve the sequence tagging model.
The priming mechanism informs the model of the conditional information by directly appending conditional information to the input text.
However, unlike \citet{DBLP:journals/corr/abs-2109-12383} that uses an integer string to represent features in a categorical style, we use a natural-language-styled indicator to better exploit the semantics of the condition. The indicators can be obtained by verbalizing the conditional information. 
\looseness=-1

Take Figure~\ref{fig:mthd_overview}(b) as an example, 
when extracting the tail-entities and the relationships for the \emph{``military''} head-entity (condition $c$), we first verbalize the entity type of \emph{``military''}, i.e., from \emph{``Org''} to \emph{``Organization''}. Then, the string \emph{``military''} and \emph{``Organization''} are appended to the input text, which serves as the information about the condition~$c$.

The priming technique leverages the self-attention mechanism in pre-trained language models and makes the token representation $z_i$ condition-aware. Hence, the representation of every $z_i$ is more \emph{task-specific} than the one in the model described in Section~\ref{sec:basic_tagging}.
More precisely, for tagging models without priming, the representation $z_i$ usually captures more general information that focuses on the context of input text. For models with priming, the representation $z_i$ is affected by the additional verbalized words when computing attention. Hence, $z_i$ becomes more task-specific and more suitable for addressing the task \cite{Zheng22wierdgpt, Zhong21pure}.
Additionally, the priming method can be easily combined with conditional features described in Section~\ref{sec:basic_tagging}.
More discussion on this will be shown in Section~\ref{sec:analysis}.

\paragraph{Relationship Priming.}
The same idea of condition priming can also be extended to relationship. 
Specifically, we decompose a relational structure extraction task into several extraction subtasks, each of them only focusing on one single relationship $r$ ($r \in \mathcal{A}$).
Similar to the condition priming, we verbalize the relationship information and append related strings to the input text as well.
Therefore, the representation $z_i$ is aware of the relationship $r$ and specific for predicting spans with relationship $r$ to the condition $c$. 

For example, in Figure~\ref{fig:mthd_overview}(c), for the given relationship \emph{``Part-Whole''}, we first verbalized it into \emph{``is part of''}. Then, the string \emph{``is part of''} is appended to the input text together with the condition priming strings. The BIO-tag sequence can be decoded into those tail-entities $\mbf s^c$ that form \emph{``Part-Whole''} relationship(s) with the given head-entity \emph{``military''}.

\paragraph{Discussion.}
A similar idea of appending tokens in the pre-trained language model's input to affect the output text representation has also been leveraged in \citet{DBLP:journals/corr/abs-2102-01373, Zhong21pure}. Yet, different from their works that only focus on relation classification and apply \emph{instance-specific} information, our \model{} with relationship priming method focuses on using \emph{task-specific information}, because we decompose relational extraction into sub-tasks. We want that different task-specific representation can be learned for different sub-tasks, hence proposing relationship priming.

An underlying advantage of \model{} with relationship priming is its ability to handle cases containing multi-relationships. After we decompose a relational structure extraction task into several extraction subtasks, we do not perform any filtering to address conflict relationship predictions between the same condition and extracted span. This is to enlarge our model's generality to different scenarios. \looseness=-1

\begin{table*}[t!]
\small
\centering
\setlength{\tabcolsep}{3.2pt}
\resizebox{.99\textwidth}{!}{
\begin{tabular}{l|ccc|ccc|ccc|ccc}
    \toprule
    \multirow{2}{*}{Model} & \multicolumn{3}{c|}{ACE05-E (en)} & \multicolumn{3}{c|}{ACE05-E (zh)} & \multicolumn{3}{c|}{ERE (en)} & \multicolumn{3}{c}{ERE (es)} \\
    & Tri-C & Arg-I & Arg-C & Tri-C & Arg-I & Arg-C & Tri-C & Arg-I & Arg-C & Tri-C & Arg-I & Arg-C \\
    \midrule
    DyGIE++$^*$ \cite{Wadden19dygiepp} 
    & 69.7 & 53.0 & 48.8 & 72.3 & 63.0 & 59.3 & 58.0 & 51.4 & 48.0 & 65.8 & 49.2 & 46.6 \\
    TANL \cite{Paolini21tanl} 
    & 68.4 & 50.1 & 47.6 & - & - & - & 54.7 & 46.6 & 43.2 & - & - & - \\
    Text2Event \cite{text2event} 
    & 71.9 & -    & 53.8 & -    & -    & -    & 59.4 & -    & 48.3 & -    & -    & -    \\
    OneIE$^*$ \cite{Lin20oneie} 
    & 74.7 & 59.2 & 56.8 & 73.3 & 63.4 & 60.5 & 57.0 & 50.1 & 46.5 & 66.5 & 54.5 & 52.2 \\
    DEGREE \cite{Hsu22degree} 
    & 73.3 & - & 55.8 & - & - & - & 57.1 & - & 49.6 & - & - & - \\
    \midrule
    \model{} w/ Cond. Priming 
    & 74.6 & \textbf{60.0} & 56.8 & 71.9 & 63.2 & 60.5 & 57.3 & 52.1 & 49.3
    & 66.3 & \textbf{55.2} & 52.6 \\
    \model{} w/ Cond. \& Rela. Priming 
    & 74.6 & 59.8 & \textbf{58.3} & 71.9 & \textbf{64.7} & \textbf{62.4}
    & 57.3 & \textbf{52.4} & \textbf{49.9}
    & 66.3 & 55.1 & \textbf{53.6}
    \\
    \bottomrule

\end{tabular}}
\caption{Results of end-to-end event extraction. All values are micro F1-score, and we highlight highest scores with boldface. \model{} with conditional and relationship priming achieves more than 1.4 Arg-C F1-score improvements in three out of four datasets. $^*$We reproduce the results using their released code. }
\label{tab:ee}
\end{table*}

\section{Experiments}
To study the effectiveness of \model{}, we consider three NLP tasks: (1) end-to-end event extraction, (2) end-to-end relation extraction, and (3) task-oriented semantic parsing. All the results are the average of five runs with different random seeds. 

\subsection{End-to-End Event Extraction}

\paragraph{Datasets.}
We consider the two most widely-used event extraction datasets, ACE-2005~\cite{Doddington04ace} and ERE~\cite{Song15ere}. For ACE-2005 (ACE05-E), we experiment on the English and Chinese portions and keep 33 event types and 22 roles, as suggested in previous works \cite{Wadden19dygiepp, Hsu22degree}. For ERE, we consider the English and Spanish annotations and follow the preprocessing of \citet{Lin20oneie} to keep 38 event types and 21 roles.

\paragraph{Baselines.}
We consider the following end-to-end event extraction models, including DyGIE++ \cite{Wadden19dygiepp}, TANL \cite{Paolini21tanl}, Text2Event \cite{text2event}, OneIE \cite{Lin20oneie}, and DEGREE \cite{Hsu22degree}.
Since \model{} requires trigger predictions, we simply take the trigger predictions made by a simple sequence tagging model trained with multi-tasking on trigger detection and named entity recognition.

For \model{}, DyGIE++, and OneIE, we consider BERT-large \cite{BERT} for ACE05-E (en) and ERE (en), and consider XLM-RoBERTa-large \cite{XLM-Roberta} for ACE05-E (zh) and ERE (es).
For generation-based models, we consider BART-large \cite{BART} for DEGREE, T5-base \cite{t5} for TANL, and T5-large \cite{t5} for Text2Event, as suggested by their original papers.

\paragraph{Implementation details.}
The followings are the training details for all baselines:

\begin{itemize}[topsep=2pt, itemsep=-2.5pt, leftmargin=13pt]
\item \textbf{DyGIE++} \cite{Wadden19dygiepp}: we use the released training script\footnote{\url{https://github.com/dwadden/dygiepp} } with the default parameters. 
\item \textbf{TANL} \cite{Paolini21tanl}: we report the numbers from the original paper.
\item \textbf{Text2Event} \cite{text2event}: we report the numbers from the original paper.
\item \textbf{OneIE} \cite{Lin20oneie}: we use the released training script\footnote{\url{http://blender.cs.illinois.edu/software/oneie/}} with the default parameters.
\item \textbf{DEGREE} \cite{Hsu22degree}: we report the numbers from the original paper.
\item \textbf{\model{}} (ours): We fine-tune pre-trained language models with the dropout rate being 0.2. We use AdamW optimizer. For parameters that are not pre-trained we set the learning rate to $10^{-3}$ and the weight decay to $10^{-3}$. For parameters that are not pre-trained we set the learning rate to $10^{-5}$ and the weight decay to $10^{-5}$. We consider the linear scheduler with a warm-up, where the warm-up epoch is 5. The number of epochs is 90. The training batch size is set to 6. For conditional token features and learnable features, the dimension is set to 100.
It takes around 6 hours to train with a NVIDIA RTX A6000 with 48GB memory.
\end{itemize}

\paragraph{Evaluation metrics.}
Following previous works \cite{Wadden19dygiepp,Lin20oneie}, we measure the correctness of arguments based on whether the offsets of the argument span match or not.
We consider argument identification F1-score (Arg-I), which cares about only the offset correctness, and argument classification F1-score (Arg-C), which cares about both offsets and the role types.
We also report trigger classification F1-score (Tri-C), although it is not our main focus as the triggers are provided via other models and we just use their predictions to simulate the end-to-end scenarios. 

\paragraph{Results.}
Table~\ref{tab:ee} shows the results of end-to-end event extraction on various datasets and languages.
Although simple, \model{} surprisingly has decent performance and achieves better results than the state-of-the-art models in terms of argument F1-scores.
We attribute the good performance to the design of priming, which leverages the semantics of the condition and makes the representations more task-specific.
It is worth noting that considering relationship priming further improves the results, which again shows the importance of task-specific representations.

\begin{table*}[t!]
\small
\centering
\setlength{\tabcolsep}{5pt}
\resizebox{.75\textwidth}{!}{
\begin{tabular}{l|ccc|ccc}
    \toprule
    \multirow{2}{*}{Model}
    & \multicolumn{3}{c|}{ACE05-R} & \multicolumn{3}{c}{ACE04-R}\\
    & Ent & Rel   & Rel+  & Ent   & Rel   & Rel+\\
    \midrule
    Table-Sequence \cite{Wang20tableseq}  
    & 89.5 & 67.6   & 64.3  & 88.6  & 63.3  & 59.6 \\
    PFN \cite{Yan21pfn}  
    & 89.0  & -     & 66.8  & 89.3  & -     & 62.5 \\
    Cascade-SRN (late fusion) \cite{Wang22Cascade}  
    & 89.4  & -     & 65.9 & -      & -     & - \\
    Cascade-SRN (early fusion) \cite{Wang22Cascade} 
    & 89.8  & -     & 67.1 & -      & -     & - \\
    PURE \cite{Zhong21pure} 
    & 89.7  & 69.0  & 65.6  & 88.8  & 64.7  & 60.2 \\
    PURE$^{\diamond}$ \cite{Zhong21pure} 
    & 90.9  & 69.4  & 67.0  & 90.3  & 66.1  & 62.2 \\
    UniRE$^{\diamond}$ \cite{Wang20unire}
    & 90.2 & - & 66.0 & 89.5 & - & \textbf{63.0} \\
    \midrule
    \model{} w/ Cond. Priming 
    & 89.6 & 69.7 & 67.3
    & 89.0 & 65.2 & 61.6
    \\
  \model{} w/ Cond. \& Rela. Priming 
    & 89.6 & \textbf{70.4} & \textbf{68.1}
    & 89.0 & \textbf{66.2} & 62.3
    \\
    \bottomrule
\end{tabular}}
\caption{Results of end-to-end relation extraction. All values are micro F1-score with the highest value in boldface. 
\model{} achieves the best performance in ACE05-R and competitive results on ACE04-R despite we get slightly lower entity scores compared to PFN. $^\diamond$indicates the use of cross-sentence context information.}
\label{tab:re}
\end{table*}

\subsection{End-to-End Relation Extraction}

\paragraph{Datasets.}
We consider two popular end-to-end relation extraction datasets, ACE04 and ACE05~\cite{Doddington04ace}, denoted as ACE04-R and ACE05-R. 
Both datasets consider 7 named entity types and 6 different relations.
We follow the same procedure in \citet{Zhong21pure} to preprocess the data and split the datasets. We refer readers to their papers for more details about the datasets.

\paragraph{Baselines.}
We compare to the following end-to-end relation extraction models: Table-Sequence \cite{Wang20tableseq}, PFN \cite{Yan21pfn}, and Cascade-SRN (both late fusion and early fusion) \cite{Wang22Cascade}. Additionally, we consider
PURE \cite{Zhong21pure}, which also takes a pipelined approach to solve end-to-end relation extraction.
To fairly compare with prior works, we use PURE's named entity predictions on the test set for \model{} to perform relational structure extraction.\footnote{We get PURE's named entity recognition predictions by retraining PURE's named entity recognition model.} In order to be consistent with our other tasks, we adopt the single sentence setting \cite{Zhong21pure} for our model. However, we also list baselines with cross-sentence settings, such as PURE's and UniRE~\cite{Wang20unire}'s results with cross-sentence context as input. 
All the models use ALBERT-xxlarge-v1~\cite{albert} as the pre-trained language models.

\paragraph{Implementation details.}
The followings are the training details for all baselines:

\begin{itemize}[topsep=2pt, itemsep=-2.5pt, leftmargin=13pt]
\item \textbf{Table-Sequence} \cite{Wang20tableseq}: we report the numbers from the original paper.
\item \textbf{Cascade-SRN} \cite{Wang22Cascade}: we report the numbers from the original paper.
\item \textbf{PURE} \cite{Zhong21pure}: we report the numbers from the original paper.
\item \textbf{PFN} \cite{Yan21pfn}: we report the numbers from the original paper.
\item \textbf{UniRE} \cite{Wang20unire}: we report the numbers from the original paper.
\item \textbf{\model{}} (ours): We fine-tune pre-trained language models with the dropout rate being 0.2. We use AdamW optimizer. For parameters that are not pre-trained we set the learning rate to $10^{-3}$ and the weight decay to $10^{-3}$. For parameters that are not pre-trained we set the learning rate to $2\times 10^{-5}$ and the weight decay to $10^{-5}$. We consider the linear scheduler with a warm-up, where the warm-up epoch is 5. The number of epochs is 30. The training batch size is set to 32. For conditional token features and learnable features, the dimension is set to 100.
It takes around 20 hours to train with a NVIDIA RTX A6000 with 48GB memory.
\end{itemize}

\paragraph{Evaluation metrics.}
We follow the standard evaluation setting with prior works~\cite{DBLP:conf/emnlp/BekoulisDDD18,Zhong21pure} and use micro F1-score as the evaluation metric. 
For named entity recognition, a predicted entity is considered as a correct prediction if its span and the entity type are both correct. We denote the score as ``Ent'' and report the scores even though it is not our main focus for evaluation.
For relation extraction, two evaluation metrics are considered: (1) Rel: a predicted relation is considered as correct when the boundaries of head-entity span and tail-entity span are correct and the predicted relation type is correct;
(2) Rel+: a stricter evaluation of Rel, where they additionally required that the entity types of head-entity span and tail-entity must also be correct.

\paragraph{Results.}
The results of end-to-end relation extraction are presented in Table~\ref{tab:re}.
From the table, we observe that \model{} has the best performance on ACE05-R and outperforms most baselines on ACE04-R.
This shows the effectiveness of \model{}.
Similar to the results of event extraction, considering relationship priming makes the representations more relationship-aware and leads to performance improvement.

\begin{table*}[t!]
\small
\centering
\setlength{\tabcolsep}{3.2pt}
\resizebox{.99\textwidth}{!}{
\begin{tabular}{l|ccc|ccc|ccc|ccc}
    \toprule
    \multirow{2}{*}{Model} & \multicolumn{3}{c|}{MTOP (en)} & \multicolumn{3}{c|}{MTOP (es)} & \multicolumn{3}{c|}{MTOP (fr)} & \multicolumn{3}{c}{MTOP (de)} \\
    & Intent & Slot-I & Slot-C & Intent & Slot-I & Slot-C & Intent & Slot-I & Slot-C & Intent & Slot-I & Slot-C \\
    \midrule
    JointBERT \cite{Li21mtop} 
    & 96.7 & -    & 92.8 & 95.2 & -    & 89.9 & 94.8 & -    & 88.3 & 95.7 & -    & 88.0 \\
    JointBERT (reproduced) 
    & 97.1 & 94.2 & 92.7 & 96.6 & 91.6 & 89.5 & 95.8 & 90.2 & 87.7 & 96.5 & 89.2 & 87.6 \\
    \midrule
    \model{} + Cond. Priming 
    & 97.1 & \textbf{94.8} & 93.4
    & 96.6 & 91.6 & 90.3 & 95.8 & \textbf{90.6} & 88.6
    & 96.5 & \textbf{89.6} & 87.9 \\
    \model{} + Cond. \& Rela. Priming 
    & 97.1 & 94.7 & \textbf{93.5}
    & 96.6 & \textbf{91.8} & \textbf{90.7}
    & 95.8 & \textbf{90.6} & \textbf{89.1}
    & 96.5 & 89.5 & \textbf{88.1} \\
    \bottomrule

\end{tabular}}
\caption{Results of task-oriented semantic parsing. Intend scores are measured in accuracy(\%) and slot scores are micro-F1 scores. The highest value is in bold.} 
\label{tab:tosp}
\end{table*}

\begin{table*}[t!]
\small
\centering
\aboverulesep = 0.2mm
\belowrulesep = 0.5mm
\setlength{\tabcolsep}{3.2pt}
\resizebox{1.0\textwidth}{!}{
\begin{tabular}{c|cc|cc|cc|cc|cc|cc|cc|cc|c}
    \toprule
    \multirow{2}{*}{Case} & 
    \multicolumn{2}{c|}{Cond.} & \multicolumn{2}{c|}{Rela.} &
    \multicolumn{2}{c|}{ACE05-E (en)} & \multicolumn{2}{c|}{ACE05-E (zh)} & 
    \multicolumn{2}{c|}{MTOP (es)} &  \multicolumn{2}{c|}{MTOP (fr)} & 
    \multicolumn{2}{c|}{ACE05-R (en)} & \multicolumn{2}{c|}{ACE04-R (en)} & \multirow{2}{*}{Average} \\
    & Feat. & Prim. & Feat. & Prim. & Arg-I & Arg-C & Arg-I & Arg-C & Slot-I & Slot-C & Slot-I & Slot-C & $\;\,$Rel$\;\,$ & Rel+ & $\;\,$Rel$\;\,$ & Rel+ & \\
    \midrule
    1 & \xmark & \xmark & \xmark & \xmark &
    57.8 & 54.2 & 60.2 & 57.2 & 91.8 & 90.2 & 90.5 & 88.4 & 67.8 & 65.5 & 62.2 & 58.9 &  69.1 \\
    2 & \cmark & \xmark & \xmark & \xmark &
    58.1 & 55.3 & 60.4 & 58.1 & \textbf{92.0} & 90.4 & 90.6 & 88.6 & 67.5 & 65.2 & 61.8 & 58.4 & 69.4 \\
    3 & \xmark & \cmark & \xmark & \xmark &
    59.6 & 56.7 & 62.0 & 59.7 & 91.8 & 90.4 & \textbf{90.7} & 88.8 & 69.6 & 67.2 & 64.7 & 60.7 & 70.6 \\
    4 & \cmark & \cmark & \xmark & \xmark &
    \textbf{60.0} & 56.8 & 63.2 & 60.5 & 91.6 & 90.3 & 90.6 & 88.7 & 69.7 & 67.3 & 65.2 & 61.6 & 70.9 \\
    \midrule
    5 & \cmark & \xmark & \cmark & \xmark &
    57.3 & 55.3 & 61.4 & 59.4 & 91.7 & 90.5 & 90.2 & 88.5 & 68.0 & 65.6 & 61.6 & 58.3 &  69.6 \\
    6 & \xmark & \cmark & \xmark & \cmark &
    59.3 & 57.6 & 63.0 & 61.2 & 91.7 & 90.5 & 90.5 & 88.9 & \textbf{70.6} & \textbf{68.2} & 66.0 & 62.2 & 71.4\\
    7 & \cmark & \cmark & \xmark & \cmark &
    59.8 & \textbf{58.3} & \textbf{64.7} & \textbf{62.4} & 91.8 & \textbf{90.7} & 90.6 & \textbf{89.1} & 70.4 & 68.1 & \textbf{66.2} & \textbf{62.3} & \textbf{71.8} \\
    8 & \cmark & \cmark & \cmark & \cmark &
    59.7 & 58.0 & 64.3 & \textbf{62.4} & 91.5 & 90.4 & 90.6 & \textbf{89.1} & 70.5 & 68.1 & 65.8 & 62.2 & 71.7 \\
    \bottomrule

\end{tabular}}
\caption{The ablation study results for three different tasks. The average column calculates the average scores of the stricter evaluation metrics (i.e, Arg-C, Slot-C, and Rel+) for each dataset. From the table, we demonstrate priming technique is the key attribute that make our sequence tagging model stronger than models with learnable features, which is the typical way of using sequence tagging models for relational structure extraction.}
\label{tab:ablation}
\end{table*}

\subsection{Task-Oriented Semantic Parsing}
\paragraph{Datasets.}
We choose MTOP \cite{Li21mtop}, a multilingual dataset on semantic parsing for task-oriented dialog systems.
We specifically consider data in English (en), Spanish (es), French (fr), and German (de).

\paragraph{Baselines.}
We consider JointBERT \cite{Chen19jointbert}, the commonly used baseline for task-oriented semantic parsing.
We directly use the predicted intents by JointBERT as the condition of \model{} for a fair comparison.
Both \model{} and JointBERT are trained with XLM-RoBERTa-large \cite{XLM-Roberta}.
Unlike event extraction and relation extraction, the condition of task-oriented semantics parsing (intent) does not include the word span, therefore, only a type feature embedding is contained in the conditional features for \model{} in this experiment.

\paragraph{Implementation details.}

The followings are the training details for all baselines:
\begin{itemize}[topsep=3pt, itemsep=-2.5pt, leftmargin=13pt]
\item \textbf{JointBERT} \cite{Chen19jointbert}: we use the training script\footnote{\url{https://github.com/monologg/JointBERT}} with the default parameters.
\item \textbf{\model{}} (ours): We fine-tune pre-trained language models with the dropout rate being 0.2. We use AdamW optimizer. For parameters that are not pre-trained we set the learning rate to $10^{-3}$ and the weight decay to $10^{-3}$. For parameters that are not pre-trained we set the learning rate to $10^{-5}$ and the weight decay to $10^{-5}$. We consider the linear scheduler with warm-up, where the warm-up epoch is 5. The number of epochs is 90. The training batch size is set to 6. For conditional token features and learnable features, the dimension is set to 100.
It takes around 4 hours to train with a NVIDIA RTX A6000 with 48GB memory.
\end{itemize}

\paragraph{Evaluation metrics.}
We following MTOP \cite{Li21mtop} to consider slot identification (Slot-I) and slot classification (Slot-C) F1-scores.
Even though we focus on the performance of slot filling, we also report the intent classification accuracy. \looseness=-1

\mypar{Results.}
As demonstrated in Table~\ref{tab:tosp}, \model{} achieves a better performance than the baselines.
Again, considering relationship priming leads to further improvement.
It is worth noting that \model{} is effective for different languages, which shows the generality of \model{}.

\subsection{Summary}

We show the superiority of \model{} on three different tasks (including ten different datasets across five different languages).
Although being a unified and simple model, the results suggest that \model{} can achieve competitive results for tasks requiring extracting relational structures.
\section{Analysis}
\label{sec:analysis}
In this section, we study two questions: (1) What is the effectiveness of priming techniques compared to learnable features? (2) Relationship priming boosts the performance of \model{}, but the task decomposition could slightly slow down the inference speed. Can we mitigate this issue?

To answer the first question, we conduct ablation experiments on sequence tagging models using different combinations of learnable features or/and adding information through the priming technique (Section~\ref{sec:ablation}). For the second question, we propose a simple modification to \model{} so that we can flexibly control the number of layers to fuse priming information to contextualized representations. The modified \model{} can serve as an efficient approximation of \model{} (Section~\ref{sec:efficient}).

\subsection{Ablation Study}
\label{sec:ablation}
We focus on the setting where we alter the choices on how to include the type information of the condition $c$ and the relationship information $r$. Table~\ref{tab:ablation} demonstrates our experimental results. 

Comparing the first four cases in Table~\ref{tab:ablation}, we observe that the addition of type features is useful in general, and using the priming technique is a more effective way to incorporate conditional information. For models in case 5 to case 8, the relationship decomposition formulation described in Section~\ref{sec:priming} is applied. Comparing case 2 to case 5, we can see that simply applying the relationship decomposition formulation for solving relational structure extraction does not lead to improvements if the way to embed the relationship $r$ is only through learnable features. However, comparing case 3 to case 6 and case 4 to case 7, we show that the relationship priming approach makes the representation $z_i$ well capture the attribute of the queried relationship, thus, better exploiting the advantage of the relationship decomposition formulation and gaining improvements. Note that we conducted preliminary experiments that use pre-trained language models' representations of the same verbalized token to be the initialization of the learnable type feature embedding, but the method shows similar results with the random initialization, hence, we stick to random initialization on the learnable type features.

\subsection{Efficient approximation of \model{}}
\label{sec:efficient}

\begin{figure}[t!]
    \centering
    \includegraphics[width=0.48\textwidth]{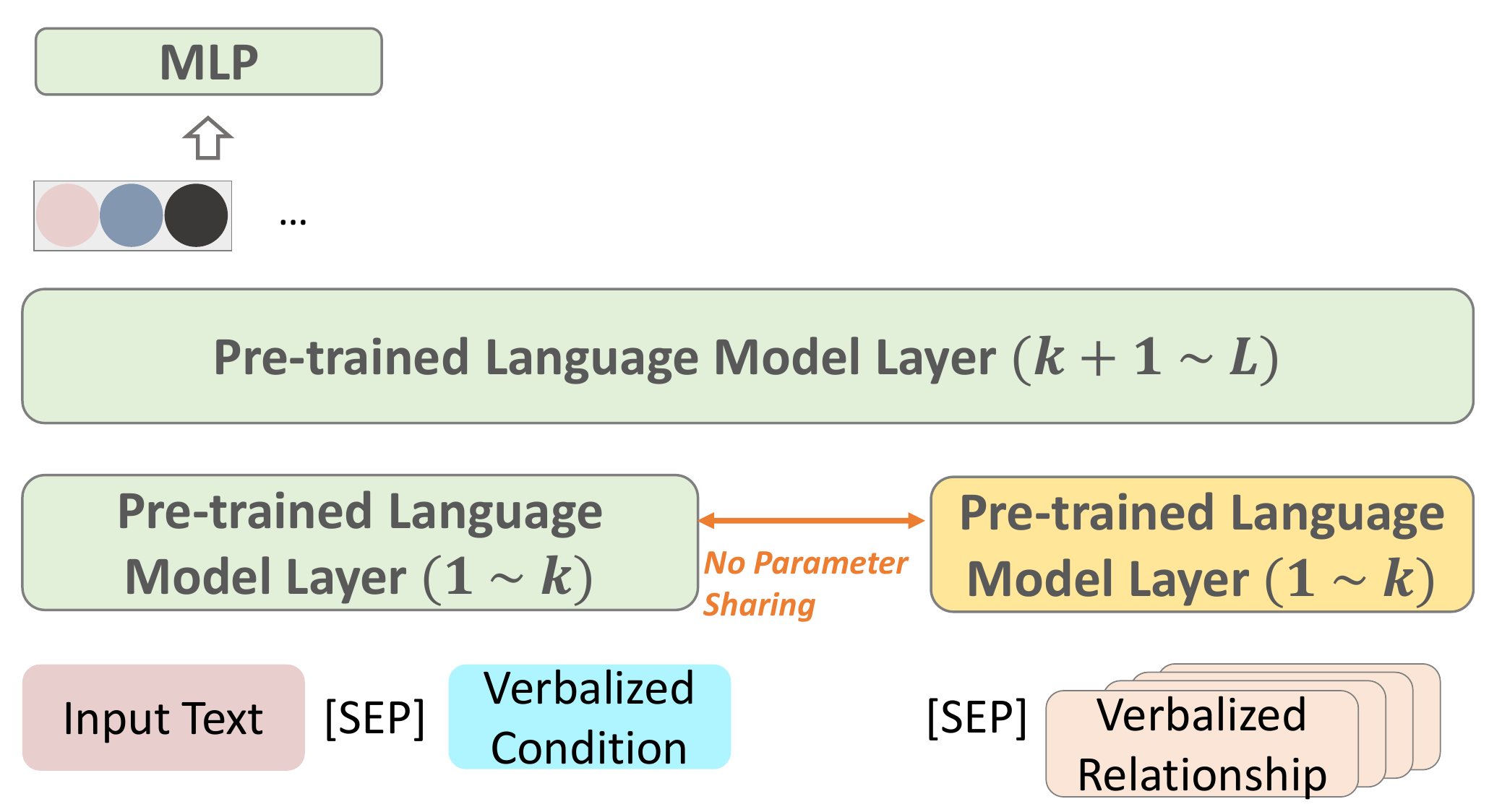}
    \caption{The illustration of our efficient approximation of \model{}, which separates the pre-trained language models into two halves to enable parallel encoding in \model{}, leading to faster inference.}
    \label{fig:cut_illustration}
\end{figure}

To make \model{} to inference faster, we perform two modifications to \model{}: (1) We first separate the pre-trained language model, which contains $L$ layers, into two halves --- one with the first $k$ layers, the other one is the remaining layers. (2) We copy the first half of the language model to another module. When an input passage is fed into the model. We use the original first half to encode the input text as well as the verbalized condition, and we use the copied first half to encode the verbalized relation. Finally, the encoded representations will be fed into the second half layers, as illustrated in Figure~\ref{fig:cut_illustration}. 
The value of $k$ is adjustable, where when $k=0$, it represents the \model{} with condition and relationship priming, and when $k=L$, it is \model{} with condition priming.

Since the encoding stage of the input text and the verbalized relationship is separated, we can accelerate the inference time of our modified \model{} through parallel encoding. More precisely, our modified \model{} can aggregate instances that share the same passage and verbalized condition. For those instances, \model{} only needs to perform the encoding once on their input passage part,\footnote{The string of verbalized relationship is usually much shorter than the input passage, hence, in most cases, the major part of the input for an instance is the input text and which requires more computations.} and paired with several separated embedded verbalized relationships, which could be parallelly encoded together.


\begin{figure}[t!]
\centering
\includegraphics[width=0.49\textwidth]{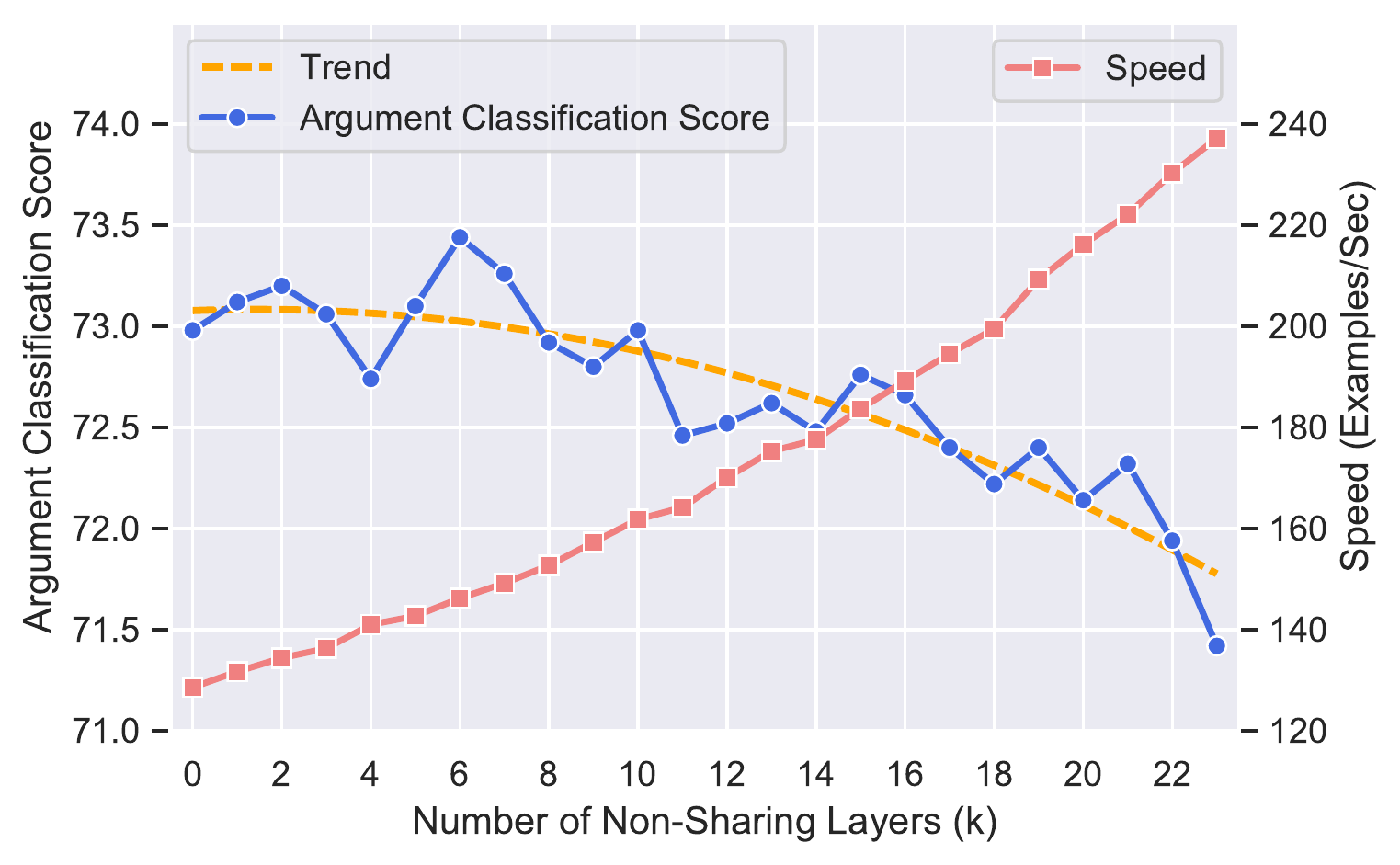}
\caption{Analysis about the performance influence and inference speed impact of our efficient approximated \model{} when the sharing layers vary. The blue line is the argument classification F1 score and the red line is the inference speed curve.}
\label{fig:cut_exp}
\end{figure}

We conduct experiments on the ACE05-E (en) dataset to test our modification. In order to better analyze the results and isolate the influence from the pipelined errors, we report the results on the event argument extraction when gold event triggers are given. The experimental results are shown in Figure~\ref{fig:cut_exp}. First, we investigate the performance influence of our modification. We find that when $k \leq 10$, the performance of our modified \model{} is strong in general and is comparable with \model{} with the condition and relationship priming. 
To compare the efficiency of the model, we benchmark the inference time by performing inference on the whole testing dataset fifty times and calculate the average speed, which is measured by checking how many instances can be processed per second. The red line in Figure~\ref{fig:cut_exp} shows the results. We observe that for our modified \model{} with $k=10$, its inference speed is around 30\% faster than the \model{} with the condition and relationship priming, but they perform similarly. 

\section{Conclusion}
\label{sec:conclusion}
In this work, we take a unified view of tasks requiring extracting relational structures and present \model{}, a simple, unified, effective, and general sequence tagging model.
The key idea is applying priming, a small trick to make the representations task-specific by appending condition-related and relationship-related strings to the input text.
Our experimental results demonstrate that \model{} is general to different tasks in various languages and can serve as a strong baseline for future research on relational structure extraction.

\section*{Acknowledgments}
We thank anonymous reviewers for their helpful feedback. 
We thank the UCLA PLUSLab and UCLA-NLP group members for the valuable discussions and comments. This research was supported in part by AFOSR MURI via Grant \#FA9550-22-1-0380, Defense Advanced Research Project Agency (DARPA) via Grant \#HR00112290103/HR0011260656, the Intelligence Advanced Research Projects Activity (IARPA) via Contract No.~2019-19051600007, National Science Foundation (NSF) via Award No.~2200274, and a research award sponsored by CISCO.
\section*{Limitations}
As we point out in Section~\ref{sec:analysis}, one of the limitations in \model{} is the inference speed. When we perform \model{} with condition and relationship priming, we requires more turns of sequence tagging processes than typical sequence tagging models. Observing this, we propose a simple way to mitigate such issue and increase the inference speed with only a small performance drop. Despite such effort, it is still slightly slower than the model requires only one pass of sequence labeling.

The other potential limitation of our method is that we assume the condition and relationship can be verbalized. However, in practice, there could be cases that the verbalization is hard to be done. Considering this, we do conduct preliminary experiments of applying \model{} with special tokens priming rather than verbalized tokens. However, our preliminary results show that such method's performance is less stable and weaker than we can achieve with \model{}.

\section*{Ethics Considerations}

\model{} fine-tunes the pre-trained language models \cite{BERT, albert}. There have been works showing the potential bias in pre-trained language models.
Although with a low possibility, especially after our finetuning, it is possible for our model to make counterfactual, and biased predictions, which may cause ethical concerns.
We suggest carefully examining those potential issues before deploying the model in any real-world applications.

\bibliographystyle{acl_natbib}
\bibliography{custom}

\begin{thebibliography}{55}
\expandafter\ifx\csname natexlab\endcsname\relax\def\natexlab#1{#1}\fi

\bibitem[{Bekoulis et~al.(2018)Bekoulis, Deleu, Demeester, and
  Develder}]{DBLP:conf/emnlp/BekoulisDDD18}
Giannis Bekoulis, Johannes Deleu, Thomas Demeester, and Chris Develder. 2018.
\newblock Adversarial training for multi-context joint entity and relation
  extraction.
\newblock In \emph{Proceedings of the 2018 Conference on Empirical Methods in
  Natural Language Processing (EMNLP)}.

\bibitem[{Chen et~al.(2019{\natexlab{a}})Chen, Chen, and
  Yu}]{chen2019incorporating}
Jiaao Chen, Jianshu Chen, and Zhou Yu. 2019{\natexlab{a}}.
\newblock Incorporating structured commonsense knowledge in story completion.
\newblock In \emph{The Thirty-Third {AAAI} Conference on Artificial
  Intelligence (AAAI)}.

\bibitem[{Chen et~al.(2019{\natexlab{b}})Chen, Zhuo, and
  Wang}]{Chen19jointbert}
Qian Chen, Zhu Zhuo, and Wen Wang. 2019{\natexlab{b}}.
\newblock {BERT} for joint intent classification and slot filling.
\newblock \emph{arXiv preprint arXiv:1902.10909}.

\bibitem[{Conneau et~al.(2020)Conneau, Khandelwal, Goyal, Chaudhary, Wenzek,
  Guzm{\'{a}}n, Grave, Ott, Zettlemoyer, and Stoyanov}]{XLM-Roberta}
Alexis Conneau, Kartikay Khandelwal, Naman Goyal, Vishrav Chaudhary, Guillaume
  Wenzek, Francisco Guzm{\'{a}}n, Edouard Grave, Myle Ott, Luke Zettlemoyer,
  and Veselin Stoyanov. 2020.
\newblock Unsupervised cross-lingual representation learning at scale.
\newblock In \emph{Proceedings of the 58th Annual Meeting of the Association
  for Computational Linguistics, {ACL}}.

\bibitem[{Desai et~al.(2021)Desai, Shrivastava, Zotov, and
  Aly}]{DBLP:journals/corr/abs-2104-07224}
Shrey Desai, Akshat Shrivastava, Alexander Zotov, and Ahmed Aly. 2021.
\newblock Low-resource task-oriented semantic parsing via intrinsic modeling.
\newblock \emph{arXiv preprint arxiv.2104.07224}.

\bibitem[{Devlin et~al.(2019)Devlin, Chang, Lee, and Toutanova}]{BERT}
Jacob Devlin, Ming{-}Wei Chang, Kenton Lee, and Kristina Toutanova. 2019.
\newblock {BERT:} pre-training of deep bidirectional transformers for language
  understanding.
\newblock In \emph{Proceedings of the 2019 Conference of the North American
  Chapter of the Association for Computational Linguistics: Human Language
  Technologies, (NAACL-HLT)}.

\bibitem[{Doddington et~al.(2004)Doddington, Mitchell, Przybocki, Ramshaw,
  Strassel, and Weischedel}]{Doddington04ace}
George~R. Doddington, Alexis Mitchell, Mark~A. Przybocki, Lance~A. Ramshaw,
  Stephanie~M. Strassel, and Ralph~M. Weischedel. 2004.
\newblock The automatic content extraction {(ACE)} program - tasks, data, and
  evaluation.
\newblock In \emph{Proceedings of the Fourth International Conference on
  Language Resources and Evaluation (LREC)}.

\bibitem[{Fincke et~al.(2022)Fincke, Agarwal, Miller, and
  Boschee}]{DBLP:journals/corr/abs-2109-12383}
Steven Fincke, Shantanu Agarwal, Scott Miller, and Elizabeth Boschee. 2022.
\newblock Language model priming for cross-lingual event extraction.
\newblock In \emph{The Thirty-Sixth {AAAI} Conference on Artificial
  Intelligence, (AAAI)}.

\bibitem[{Fu et~al.(2019)Fu, Li, and Ma}]{DBLP:conf/acl/FuLM19}
Tsu{-}Jui Fu, Peng{-}Hsuan Li, and Wei{-}Yun Ma. 2019.
\newblock Graphrel: Modeling text as relational graphs for joint entity and
  relation extraction.
\newblock In \emph{Proceedings of the 57th Conference of the Association for
  Computational Linguistics (ACL)}.

\bibitem[{Fusco et~al.(2022)Fusco, Pascual, and Staar}]{fusco2022mixer}
Francesco Fusco, Damian Pascual, and Peter Staar. 2022.
\newblock pnlp-mixer: an efficient all-mlp architecture for language.
\newblock \emph{arXiv preprint arxiv.2202.04350}.

\bibitem[{Gupta et~al.(2018)Gupta, Shah, Mohit, Kumar, and Lewis}]{Gupta18top}
Sonal Gupta, Rushin Shah, Mrinal Mohit, Anuj Kumar, and Mike Lewis. 2018.
\newblock Semantic parsing for task oriented dialog using hierarchical
  representations.
\newblock In \emph{Proceedings of the 2018 Conference on Empirical Methods in
  Natural Language Processing (EMNLP)}.

\bibitem[{Han et~al.(2019)Han, Hsu, Yang, Galstyan, Weischedel, and
  Peng}]{han-etal-2019-deep}
Rujun Han, I-Hung Hsu, Mu~Yang, Aram Galstyan, Ralph Weischedel, and Nanyun
  Peng. 2019.
\newblock Deep structured neural network for event temporal relation
  extraction.
\newblock In \emph{Proceedings of the 23rd Conference on Computational Natural
  Language Learning (CoNLL)}.

\bibitem[{Hsu et~al.(2022{\natexlab{a}})Hsu, Guo, Natarajan, and
  Peng}]{hsu2021discourse}
I-Hung Hsu, Xiao Guo, Premkumar Natarajan, and Nanyun Peng. 2022{\natexlab{a}}.
\newblock Discourse-level relation extraction via graph pooling.
\newblock In \emph{The Thirty-Sixth AAAI Conference On Artificial Intelligence
  Workshop on Deep Learning on Graphs: Method and Applications (DLG-AAAI)}.

\bibitem[{Hsu et~al.(2022{\natexlab{b}})Hsu, Huang, Boschee, Miller, Natarajan,
  Chang, and Peng}]{Hsu22degree}
I{-}Hung Hsu, Kuan{-}Hao Huang, Elizabeth Boschee, Scott Miller, Prem
  Natarajan, Kai{-}Wei Chang, and Nanyun Peng. 2022{\natexlab{b}}.
\newblock Degree: A data-efficient generation-based event extraction model.
\newblock In \emph{Proceedings of the 2022 Conference of the North American
  Chapter of the Association for Computational Linguistics: Human Language
  Technologies (NAACL-HLT)}.

\bibitem[{Huang et~al.(2022)Huang, Hsu, Natarajan, Chang, and
  Peng}]{acl2022xgear}
Kuan-Hao Huang, I-Hung Hsu, Premkumar Natarajan, Kai-Wei Chang, and Nanyun
  Peng. 2022.
\newblock Multilingual generative language models for zero-shot cross-lingual
  event argument extraction.
\newblock In \emph{Proceedings of the 60th Annual Meeting of the Association
  for Computational Linguistics (ACL)}.

\bibitem[{Huang and Peng(2021)}]{huang2021document}
Kung-Hsiang Huang and Nanyun Peng. 2021.
\newblock Document-level event extraction with efficient end-to-end learning of
  cross-event dependencies.
\newblock In \emph{The 3rd Workshop on Narrative Understanding (NAACL 2021)}.

\bibitem[{Katiyar and Cardie(2017)}]{DBLP:conf/acl/KatiyarC17}
Arzoo Katiyar and Claire Cardie. 2017.
\newblock Going out on a limb: Joint extraction of entity mentions and
  relations without dependency trees.
\newblock In \emph{Proceedings of the 55th Annual Meeting of the Association
  for Computational Linguistics, {ACL}}.

\bibitem[{Lafferty et~al.(2001)Lafferty, McCallum, and
  Pereira}]{DBLP:conf/icml/LaffertyMP01}
John~D. Lafferty, Andrew McCallum, and Fernando C.~N. Pereira. 2001.
\newblock Conditional random fields: Probabilistic models for segmenting and
  labeling sequence data.
\newblock In \emph{Proceedings of the Eighteenth International Conference on
  Machine Learning (ICML)}.

\bibitem[{Lan et~al.(2020)Lan, Chen, Goodman, Gimpel, Sharma, and
  Soricut}]{albert}
Zhenzhong Lan, Mingda Chen, Sebastian Goodman, Kevin Gimpel, Piyush Sharma, and
  Radu Soricut. 2020.
\newblock {ALBERT:} {A} lite {BERT} for self-supervised learning of language
  representations.
\newblock In \emph{8th International Conference on Learning Representations
  (ICLR)}.

\bibitem[{Lewis et~al.(2020)Lewis, Liu, Goyal, Ghazvininejad, Mohamed, Levy,
  Stoyanov, and Zettlemoyer}]{BART}
Mike Lewis, Yinhan Liu, Naman Goyal, Marjan Ghazvininejad, Abdelrahman Mohamed,
  Omer Levy, Veselin Stoyanov, and Luke Zettlemoyer. 2020.
\newblock {BART:} denoising sequence-to-sequence pre-training for natural
  language generation, translation, and comprehension.
\newblock In \emph{Proceedings of the 58th Annual Meeting of the Association
  for Computational Linguistics (ACL)}.

\bibitem[{Li et~al.(2021)Li, Arora, Chen, Gupta, Gupta, and Mehdad}]{Li21mtop}
Haoran Li, Abhinav Arora, Shuohui Chen, Anchit Gupta, Sonal Gupta, and Yashar
  Mehdad. 2021.
\newblock {MTOP:} {A} comprehensive multilingual task-oriented semantic parsing
  benchmark.
\newblock In \emph{Proceedings of the 16th Conference of the European Chapter
  of the Association for Computational Linguistics (EACL)}.

\bibitem[{Li et~al.(2013)Li, Ji, and Huang}]{Li13jointbeam}
Qi~Li, Heng Ji, and Liang Huang. 2013.
\newblock Joint event extraction via structured prediction with global
  features.
\newblock In \emph{Proceedings of the 51st Annual Meeting of the Association
  for Computational Linguistics (ACL)}.

\bibitem[{Lin et~al.(2020)Lin, Ji, Huang, and Wu}]{Lin20oneie}
Ying Lin, Heng Ji, Fei Huang, and Lingfei Wu. 2020.
\newblock A joint neural model for information extraction with global features.
\newblock In \emph{Proceedings of the 58th Annual Meeting of the Association
  for Computational Linguistics (ACL)}.

\bibitem[{Liu et~al.(2018)Liu, Chen, Ren, Feng, Liu, and
  Yin}]{liu2018knowledge}
Shuman Liu, Hongshen Chen, ZhaDBLP:conf/acl/LiuFCRYL18ochun Ren, Yang Feng, Qun
  Liu, and Dawei Yin. 2018.
\newblock Knowledge diffusion for neural dialogue generation.
\newblock In \emph{Proceedings of the 56th Annual Meeting of the Association
  for Computational Linguistics (ACL)}.

\bibitem[{Louvan and Magnini(2020)}]{DBLP:conf/coling/LouvanM20}
Samuel Louvan and Bernardo Magnini. 2020.
\newblock Recent neural methods on slot filling and intent classification for
  task-oriented dialogue systems: {A} survey.
\newblock In \emph{Proceedings of the 28th International Conference on
  Computational Linguistics (COLING)}.

\bibitem[{Lu et~al.(2022)Lu, Hsu, Zhou, Ma, and Chen}]{lu2022Summ}
Keming Lu, I-Hung Hsu, Wenxuan Zhou, Mingyu~Derek Ma, and Muhao Chen. 2022.
\newblock Summarization as indirect supervision for relation extraction.
\newblock \emph{arXiv preprint arXiv:2205.09837}.

\bibitem[{Lu et~al.(2021)Lu, Lin, Xu, Han, Tang, Li, Sun, Liao, and
  Chen}]{text2event}
Yaojie Lu, Hongyu Lin, Jin Xu, Xianpei Han, Jialong Tang, Annan Li, Le~Sun,
  Meng Liao, and Shaoyi Chen. 2021.
\newblock Text2event: Controllable sequence-to-structure generation for
  end-to-end event extraction.
\newblock In \emph{Proceedings of the 59th Annual Meeting of the Association
  for Computational Linguistics and the 11th International Joint Conference on
  Natural Language Processing (ACL/IJCNLP)}.

\bibitem[{Lyu and Chen(2021)}]{lyu-chen-2021-relation}
Shengfei Lyu and Huanhuan Chen. 2021.
\newblock Relation classification with entity type restriction.
\newblock In \emph{Findings of the Association for Computational Linguistics:
  ACL-IJCNLP 2021}.

\bibitem[{Ma et~al.(2020)Ma, Wang, Anubhai, Ballesteros, and
  Al{-}Onaizan}]{DBLP:conf/emnlp/MaWABA20}
Jie Ma, Shuai Wang, Rishita Anubhai, Miguel Ballesteros, and Yaser
  Al{-}Onaizan. 2020.
\newblock Resource-enhanced neural model for event argument extraction.
\newblock In \emph{Findings of the Association for Computational Linguistics
  (EMNLP-Findings)}.

\bibitem[{Miwa and Bansal(2016)}]{DBLP:conf/acl/MiwaB16}
Makoto Miwa and Mohit Bansal. 2016.
\newblock End-to-end relation extraction using lstms on sequences and tree
  structures.
\newblock In \emph{Proceedings of the 54th Annual Meeting of the Association
  for Computational Linguistics, {ACL}}.

\bibitem[{Nguyen et~al.(2016)Nguyen, Cho, and Grishman}]{Nguyen16jrnn}
Thien~Huu Nguyen, Kyunghyun Cho, and Ralph Grishman. 2016.
\newblock Joint event extraction via recurrent neural networks.
\newblock In \emph{The 2016 Conference of the North American Chapter of the
  Association for Computational Linguistics: Human Language Technologies
  (NAACL-HLT)}.

\bibitem[{Nguyen and Grishman(2015)}]{nguyen-grishman-2015-event}
Thien~Huu Nguyen and Ralph Grishman. 2015.
\newblock Event detection and domain adaptation with convolutional neural
  networks.
\newblock In \emph{Proceedings of the 53rd Annual Meeting of the Association
  for Computational Linguistics (ACL)}.

\bibitem[{Paolini et~al.(2021)Paolini, Athiwaratkun, Krone, Ma, Achille,
  Anubhai, dos Santos, Xiang, and Soatto}]{Paolini21tanl}
Giovanni Paolini, Ben Athiwaratkun, Jason Krone, Jie Ma, Alessandro Achille,
  Rishita Anubhai, C{\'{\i}}cero~Nogueira dos Santos, Bing Xiang, and Stefano
  Soatto. 2021.
\newblock Structured prediction as translation between augmented natural
  languages.
\newblock In \emph{9th International Conference on Learning Representations
  (ICLR)}.

\bibitem[{Peng et~al.(2020)Peng, Gao, Han, Lin, Li, Liu, Sun, and
  Zhou}]{peng-etal-2020-learning}
Hao Peng, Tianyu Gao, Xu~Han, Yankai Lin, Peng Li, Zhiyuan Liu, Maosong Sun,
  and Jie Zhou. 2020.
\newblock {L}earning from {C}ontext or {N}ames? {A}n {E}mpirical {S}tudy on
  {N}eural {R}elation {E}xtraction.
\newblock In \emph{Proceedings of the 2020 Conference on Empirical Methods in
  Natural Language Processing (EMNLP)}.

\bibitem[{Raffel et~al.(2020)Raffel, Shazeer, Roberts, Lee, Narang, Matena,
  Zhou, Li, and Liu}]{t5}
Colin Raffel, Noam Shazeer, Adam Roberts, Katherine Lee, Sharan Narang, Michael
  Matena, Yanqi Zhou, Wei Li, and Peter~J. Liu. 2020.
\newblock Exploring the limits of transfer learning with a unified text-to-text
  transformer.
\newblock \emph{J. Mach. Learn. Res.}, 21:140:1--140:67.

\bibitem[{Song et~al.(2015)Song, Bies, Strassel, Riese, Mott, Ellis, Wright,
  Kulick, Ryant, and Ma}]{Song15ere}
Zhiyi Song, Ann Bies, Stephanie~M. Strassel, Tom Riese, Justin Mott, Joe Ellis,
  Jonathan Wright, Seth Kulick, Neville Ryant, and Xiaoyi Ma. 2015.
\newblock From light to rich {ERE:} annotation of entities, relations, and
  events.
\newblock In \emph{Proceedings of the The 3rd Workshop on {EVENTS:} Definition,
  Detection, Coreference, and Representation, (EVENTS@HLP-NAACL)}.

\bibitem[{Sun et~al.(2019)Sun, Gong, Wu, Gong, Jiang, Lan, Sun, and
  Duan}]{DBLP:conf/acl/SunGWGJLSD19}
Changzhi Sun, Yeyun Gong, Yuanbin Wu, Ming Gong, Daxin Jiang, Man Lan, Shiliang
  Sun, and Nan Duan. 2019.
\newblock Joint type inference on entities and relations via graph
  convolutional networks.
\newblock In \emph{Proceedings of the 57th Conference of the Association for
  Computational Linguistics (ACL)}.

\bibitem[{T{\"{u}}r et~al.(2010)T{\"{u}}r, Hakkani{-}T{\"{u}}r, and
  Heck}]{Tur10atis}
G{\"{o}}khan T{\"{u}}r, Dilek Hakkani{-}T{\"{u}}r, and Larry~P. Heck. 2010.
\newblock What is left to be understood in atis?
\newblock In \emph{2010 {IEEE} Spoken Language Technology Workshop (SLT)}.

\bibitem[{Wadden et~al.(2019)Wadden, Wennberg, Luan, and
  Hajishirzi}]{Wadden19dygiepp}
David Wadden, Ulme Wennberg, Yi~Luan, and Hannaneh Hajishirzi. 2019.
\newblock Entity, relation, and event extraction with contextualized span
  representations.
\newblock In \emph{Proceedings of the 2019 Conference on Empirical Methods in
  Natural Language Processing and the 9th International Joint Conference on
  Natural Language Processing (EMNLP-IJCNLP)}.

\bibitem[{Wang et~al.(2022)Wang, Liu, Le, and Yokota}]{Wang22Cascade}
An~Wang, Ao~Liu, Hieu~Hanh Le, and Haruo Yokota. 2022.
\newblock Towards effective multi-task interaction for entity-relation
  extraction: {A} unified framework with selection recurrent network.
\newblock \emph{arXiv preprint arXiv:2202.07281}.

\bibitem[{Wang and Lu(2020)}]{Wang20tableseq}
Jue Wang and Wei Lu. 2020.
\newblock Two are better than one: Joint entity and relation extraction with
  table-sequence encoders.
\newblock In \emph{Proceedings of the 2020 Conference on Empirical Methods in
  Natural Language Processing (EMNLP)}.

\bibitem[{Wang et~al.(2019)Wang, Wang, Han, Liu, Li, Li, Sun, Zhou, and
  Ren}]{wang-etal-2019-hmeae}
Xiaozhi Wang, Ziqi Wang, Xu~Han, Zhiyuan Liu, Juanzi Li, Peng Li, Maosong Sun,
  Jie Zhou, and Xiang Ren. 2019.
\newblock {HMEAE:} hierarchical modular event argument extraction.
\newblock In \emph{Proceedings of the 2019 Conference on Empirical Methods in
  Natural Language Processing and the 9th International Joint Conference on
  Natural Language Processing (EMNLP-IJCNLP)}.

\bibitem[{Wang et~al.(2021)Wang, Sun, Wu, Zhou, Li, and Yan}]{Wang20unire}
Yijun Wang, Changzhi Sun, Yuanbin Wu, Hao Zhou, Lei Li, and Junchi Yan. 2021.
\newblock Unire: {A} unified label space for entity relation extraction.
\newblock In \emph{Proceedings of the 59th Annual Meeting of the Association
  for Computational Linguistics and the 11th International Joint Conference on
  Natural Language Processing (ACL/IJCNLP)}.

\bibitem[{Wei et~al.(2020)Wei, Su, Wang, Tian, and
  Chang}]{DBLP:conf/acl/WeiSWTC20}
Zhepei Wei, Jianlin Su, Yue Wang, Yuan Tian, and Yi~Chang. 2020.
\newblock A novel cascade binary tagging framework for relational triple
  extraction.
\newblock In \emph{Proceedings of the 58th Annual Meeting of the Association
  for Computational Linguistics (ACL)}.

\bibitem[{Wu and He(2019)}]{wu2019enriching}
Shanchan Wu and Yifan He. 2019.
\newblock Enriching pre-trained language model with entity information for
  relation classification.
\newblock In \emph{Proceedings of the 28th ACM international conference on
  information and knowledge management}, pages 2361--2364.

\bibitem[{Yan et~al.(2021)Yan, Zhang, Fu, Zhang, and Wei}]{Yan21pfn}
Zhiheng Yan, Chong Zhang, Jinlan Fu, Qi~Zhang, and Zhongyu Wei. 2021.
\newblock A partition filter network for joint entity and relation extraction.
\newblock In \emph{Proceedings of the 2021 Conference on Empirical Methods in
  Natural Language Processing (EMNLP)}.

\bibitem[{Yang et~al.(2019)Yang, Feng, Qiao, Kan, and
  Li}]{DBLP:conf/acl/YangFQKL19}
Sen Yang, Dawei Feng, Linbo Qiao, Zhigang Kan, and Dongsheng Li. 2019.
\newblock Exploring pre-trained language models for event extraction and
  generation.
\newblock In \emph{Proceedings of the 57th Conference of the Association for
  Computational Linguistics (ACL)}.

\bibitem[{Yasunaga et~al.(2021)Yasunaga, Ren, Bosselut, Liang, and
  Leskovec}]{yasunaga-etal-2021-qa}
Michihiro Yasunaga, Hongyu Ren, Antoine Bosselut, Percy Liang, and Jure
  Leskovec. 2021.
\newblock {QA}-{GNN}: Reasoning with language models and knowledge graphs for
  question answering.
\newblock In \emph{Proceedings of the 2021 Conference of the North American
  Chapter of the Association for Computational Linguistics: Human Language
  Technologies}.

\bibitem[{Yu et~al.(2020)Yu, Zhang, Shu, Liu, Wang, Wang, and
  Li}]{DBLP:conf/ecai/0002ZSLWWL20}
Bowen Yu, Zhenyu Zhang, Xiaobo Shu, Tingwen Liu, Yubin Wang, Bin Wang, and
  Sujian Li. 2020.
\newblock Joint extraction of entities and relations based on a novel
  decomposition strategy.
\newblock In \emph{24th European Conference on Artificial Intelligence (ECAI)}.

\bibitem[{Zhang et~al.(2018)Zhang, Fang, Cao, Liu, Chen, and
  Tan}]{DBLP:conf/icann/ZhangFCLCT18}
Dongjie Zhang, Zheng Fang, Yanan Cao, Yanbing Liu, Xiaojun Chen, and Jianlong
  Tan. 2018.
\newblock Attention-based {RNN} model for joint extraction of intent and word
  slot based on a tagging strategy.
\newblock In \emph{Artificial Neural Networks and Machine Learning (ICANN)}.

\bibitem[{Zheng and Lapata(2022)}]{Zheng22wierdgpt}
Hao Zheng and Mirella Lapata. 2022.
\newblock Disentangled sequence to sequence learning for compositional
  generalization.
\newblock In \emph{Proceedings of the 60th Annual Meeting of the Association
  for Computational Linguistics (ACL)}.

\bibitem[{Zheng et~al.(2017)Zheng, Wang, Bao, Hao, Zhou, and
  Xu}]{DBLP:conf/acl/ZhengWBHZX17}
Suncong Zheng, Feng Wang, Hongyun Bao, Yuexing Hao, Peng Zhou, and Bo~Xu. 2017.
\newblock Joint extraction of entities and relations based on a novel tagging
  scheme.
\newblock In \emph{Proceedings of the 55th Annual Meeting of the Association
  for Computational Linguistics, {ACL} 2017}.

\bibitem[{Zhong and Chen(2021)}]{Zhong21pure}
Zexuan Zhong and Danqi Chen. 2021.
\newblock A frustratingly easy approach for entity and relation extraction.
\newblock In \emph{Proceedings of the 2021 Conference of the North American
  Chapter of the Association for Computational Linguistics: Human Language
  Technologies (NAACL-HLT)}.

\bibitem[{Zhou and Chen(2021{\natexlab{a}})}]{zhou2021improved}
Wenxuan Zhou and Muhao Chen. 2021{\natexlab{a}}.
\newblock An improved baseline for sentence-level relation extraction.
\newblock \emph{arXiv preprint arXiv:2102.01373}.

\bibitem[{Zhou and
  Chen(2021{\natexlab{b}})}]{DBLP:journals/corr/abs-2102-01373}
Wenxuan Zhou and Muhao Chen. 2021{\natexlab{b}}.
\newblock An improved baseline for sentence-level relation extraction.
\newblock \emph{arXiv preprint arXiv:2102.0137}.

\end{thebibliography}

\clearpage
\appendix

\section{Detailed Results}
Table~\ref{tab:detail_ee}, \ref{tab:detail_re}, and \ref{tab:detail_tosp} lists the detailed results (mean  and standard deviation) of \model{}.

\begin{table}[ht!]
\begin{minipage}{.99\textwidth}
\small
\centering
\setlength{\tabcolsep}{3pt}
\resizebox{.99\textwidth}{!}{
\begin{tabular}{l|cc|cc|cc|cc}
    \toprule
    \multirow{2}{*}{Model} & \multicolumn{2}{c|}{ACE05-E (en)} & \multicolumn{2}{c|}{ACE05-E (zh)} & \multicolumn{2}{c|}{ERE (en)} & \multicolumn{2}{c}{ERE (es)} \\
    & Arg-I & Arg-C & Arg-I & Arg-C & Arg-I & Arg-C & Arg-I & Arg-C \\
    \midrule
    DyGIE++$^*$ \cite{Wadden19dygiepp} 
    & 53.0 & 48.8 & 63.0 & 59.3 & 51.4 & 48.0 & 49.2 & 46.6 \\
    TANL \cite{Paolini21tanl} 
    & 50.1 & 47.6 & - & - & 46.6 & 43.2 & - & - \\
    Text2Event \cite{text2event} 
    & -    & 53.8 & -    & -    & -    & 48.3 & -    & -    \\
    OneIE$^*$ \cite{Lin20oneie} 
    & 59.2 & 56.8 & 63.4 & 60.5 & 50.1 & 46.5 & 54.5 & 52.2 \\
    DEGREE \cite{Hsu22degree} 
    & - & 55.8 & - & - & - & 49.6 & - & - \\
    \midrule
    \model{} w/ Cond. Priming 
    & \textbf{60.0}{\tiny$\pm 0.47$} & 56.8{\tiny$\pm 0.54$} & 63.2{\tiny$\pm 0.74$} & 60.5{\tiny$\pm 0.73$} & 52.1{\tiny$\pm 0.15$} & 49.3{\tiny$\pm 0.28$} & \textbf{55.2}{\tiny$\pm 0.79$} & 52.6{\tiny$\pm 1.11$} \\
    \model{} w/ Cond. \& Rel. Priming 
    &   59.8{\tiny$\pm 0.53$} & \textbf{58.3}{\tiny$\pm 0.67$} & \textbf{64.7}{\tiny$\pm 0.88$} & \textbf{62.4}{\tiny$\pm 0.85$} & \textbf{52.4}{\tiny$\pm 0.41$} & \textbf{49.9}{\tiny$\pm 0.60$} & 55.1{\tiny$\pm 0.89$} & \textbf{53.6}{\tiny$\pm 0.83$} \\
    \bottomrule

\end{tabular}}

\caption{Detailed results of end-to-end event extraction (mean$\pm$std). All values are micro F1-score, and we highlight the highest scores with boldface. $^*$We reproduce the results using their released code.}
\label{tab:detail_ee}

\end{minipage}
\end{table}

\begin{table}[ht!]
\begin{minipage}{.99\textwidth}
\small
\centering
\setlength{\tabcolsep}{5pt}
\resizebox{.75\textwidth}{!}{
\begin{tabular}{l|cc|cc}
    \toprule
    \multirow{2}{*}{Model}
    & \multicolumn{2}{c|}{ACE05-R} & \multicolumn{2}{c}{ACE04-R}\\
    & Rel   & Rel+    & Rel   & Rel+\\
    \midrule
    Table-Sequence \cite{Wang20tableseq}  
    & 67.6   & 64.3  & 63.3  & 59.6 \\
    PFN \cite{Yan21pfn}  
    & -     & 66.8  & -     & 62.5 \\
    Cascade-SRN (late fusion) \cite{Wang22Cascade}  
    & -     & 65.9 & -     & - \\
    Cascade-SRN (early fusion) \cite{Wang22Cascade}
    & -     & 67.1 & -     & - \\
    PURE \cite{Zhong21pure} 
    & 69.0  & 65.6  & 64.7  & 60.2 \\
    PURE$^{\diamond}$ \cite{Zhong21pure} 
    & 69.4  & 67.0  & 66.1  & 62.2 \\
    UniRE$^{\diamond}$ \cite{Wang20unire}
    & - & 66.0 &  - & \textbf{63.0} \\
    \midrule
    \model{} w/ Cond. Priming 
    & 69.7{\tiny$\pm 0.73$} & 67.3{\tiny$\pm 0.61$} & 65.2{\tiny$\pm 1.56$} & 61.6{\tiny$\pm 1.65$} \\
   \model{} w/ Cond. \& Rela. Priming
    & \textbf{70.4}{\tiny$\pm 0.64$} & \textbf{68.1}{\tiny$\pm 0.64$} & \textbf{66.2}{\tiny$\pm 1.51$} & 62.3{\tiny$\pm 1.19$} \\
    \bottomrule
\end{tabular}}
\caption{Detailed results of end-to-end relation extraction (mean$\pm$std). All values are micro F1-score with the highest value in bold. Note that in ACE04-R, the experiment was conducted and evaluated through 5-fold cross-validation, hence the variance is slightly larger compared to ACE05-R, which fixes the test set for every run with a different random seed. $^\diamond$indicates the use of cross-sentence context information.}
\label{tab:detail_re}
\end{minipage}
\end{table}

\begin{table}[ht!]
\begin{minipage}{.99\textwidth}
\small
\centering
\setlength{\tabcolsep}{3.2pt}
\resizebox{.99\textwidth}{!}{
\begin{tabular}{l|cc|cc|cc|cc}
    \toprule
    \multirow{2}{*}{Model} & \multicolumn{2}{c|}{MTOP (en)} & \multicolumn{2}{c|}{MTOP (es)} & \multicolumn{2}{c|}{MTOP (fr)} & \multicolumn{2}{c}{MTOP (de)} \\
    & Slot-I & Slot-C & Slot-I & Slot-C & Slot-I & Slot-C & Slot-I & Slot-C \\
    \midrule
    JointBERT \cite{Li21mtop} 
    & -    & 92.8 & -    & 89.9 & -    & 88.3 & -    & 88.0 \\
    JointBERT (reproduced) 
    & 94.2 & 92.7 & 91.6 & 89.5 & 90.2 & 87.7 & 89.2 & 87.6 \\
    \midrule
    \model{} + Cond. Priming 
    & \textbf{94.8}{\tiny$\pm 0.27$} & 93.4{\tiny$\pm 0.30$} & 91.6{\tiny$\pm 0.43$} & 90.3{\tiny$\pm 0.15$} & \textbf{90.6}{\tiny$\pm 0.22$} & 88.6{\tiny$\pm 0.24$} & \textbf{89.6}{\tiny$\pm 0.15$} & 87.9{\tiny$\pm 0.07$} \\
    \model{} + Cond. \& Rela. Priming 
    & 94.7{\tiny$\pm 0.07$} & \textbf{93.5}{\tiny$\pm 0.13$} & \textbf{91.8}{\tiny$\pm 0.16$} & \textbf{90.7}{\tiny$\pm 0.14$} & \textbf{90.6}{\tiny$\pm 0.36$} & \textbf{89.1}{\tiny$\pm 0.35$} & 89.5{\tiny$\pm 0.34$} & \textbf{88.1}{\tiny$\pm 0.36$} \\
    \bottomrule

\end{tabular}}
\caption{Detailed results of task-oriented semantic parsing (mean$\pm$std). Intend scores are measured in accuracy(\%) and slot scores are micro-F1 scores. The highest value is in bold.}
\label{tab:detail_tosp}
\end{minipage}
\end{table}



\end{document}